\documentclass[twocolumn,12pt]{article}
\usepackage[utf8]{inputenc}

\usepackage{hyperref}
\usepackage{amsmath}

\usepackage{lipsum}
\usepackage[switch]{lineno}
\usepackage[svgnames]{xcolor}

\usepackage{tikz}
\usetikzlibrary{tikzmark}
\usepackage[most]{tcolorbox}

\tcbset{on line, 
        boxsep=6pt, left=-2pt,right=-4pt,top=0pt,bottom=0pt,  
        highlight math style={enhanced}
        }

\newcommand{\eq}{\begin{equation}}
\newcommand{\eqx}{\end{equation}}
\newcommand{\eqn}{\begin{eqnarray}}
\newcommand{\eqnx}{\end{eqnarray}}

\newcommand{\exampletext}[1]{%
\bigskip
\noindent{}\textit{#1}\\
}

\title{Aspects of human memory and Large Language Models}

\author{Romuald A. Janik\thanks{e-mail: {\tt romuald.janik@gmail.com}}
\\ \\ 
\small 
Institute of Theoretical Physics and Mark Kac Center for Complex Systems Research\\\small
Jagiellonian University, ul. {\L}ojasiewicza 11, 30-348 Krak{\'o}w, Poland}
\date{}

\usepackage{fullpage}
\usepackage{graphicx}

\begin{document}

\maketitle

\begin{abstract}
Large Language Models (LLMs) are huge artificial neural networks which primarily serve to generate text, but also provide a very sophisticated probabilistic model of language use. Since generating a semantically consistent text requires a form of effective memory, we investigate the memory properties of LLMs and find surprising similarities with key characteristics of human memory. We argue that the human-like memory properties of the Large Language Model do not follow automatically from the LLM architecture but are rather learned from the statistics of the training textual data. These results strongly suggest that the biological features of human memory leave an imprint on the way that we structure our textual narratives.
\end{abstract}

\section{Introduction}

The use of language is a hallmark of \textit{Homo Sapiens}, and can be viewed as a crucial driving force of the immense advancement of our species in the last thousands of years. 
Language is understood here at the most general level, comprising not only grammatical rules, but also semantics and the global structure of narratives, all necessary for it to perform the function of a complex means of communication.

\begin{figure}[t!]
\begin{center}
\includegraphics[width=0.45\textwidth]{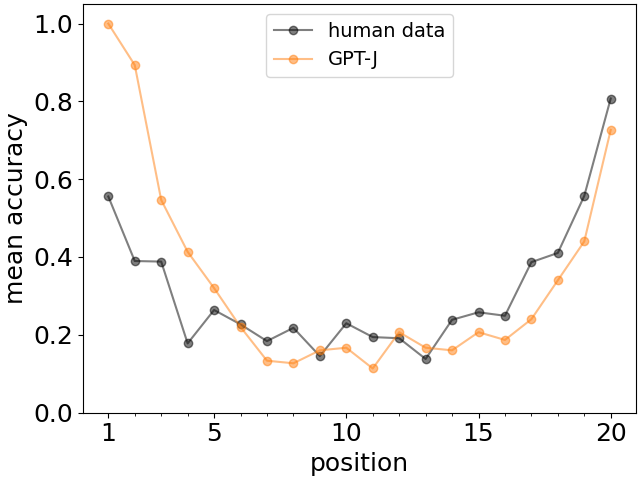}
\end{center}
  \caption{Recall accuracy for a serial memory experiment with human subjects (sample data from~\cite{glanzercunitz}) and for a memorization experiment of a list of 20 facts of the \textit{has-a} type for
  the Large Language Model \texttt{GPT-J}~\cite{gptj} studied extensively in this paper.
  %two Large Language Models of similar size (\texttt{GPT-J}~\cite{gptj} is the model studied extensively in this paper). 
  The observed U-shaped curves exhibit the \emph{primacy} and \emph{recency} effects.}
  \label{fig.human}
\end{figure}

Naturally, it is a very interesting question to understand the interrelationships between human cognitive abilities
%, and thus the properties of the biological brain, 
and the properties of language conceptualized in the above very general sense. 
%We argue that Large Language Models (LLM) can provide some data going in this direction.
We argue that Large Language Models~\cite{llm,transformer} (LLMs) can be employed as a very useful tool in investigations going in this direction.
As we explain in the following section, LLMs provide for us a very sophisticated probabilistic model of language use, which is extracted from immense corpora of text.

We focus on perhaps the simplest cognitive prerequisite for making a coherent narrative -- memory. Indeed, when writing or generating text, one has to keep track of facts already presented so as not to contradict them later in the text. 
From the above perspective, the study of memory is especially interesting as it would seem that the specific characteristics of memory should not really matter much for preserving the coherence of the generated text. 
Yet it is well known that despite its apparent simplicity, human memory exhibits some quite particular properties,
which
have been studied by cognitive psychologists for over a century. Human memory manifests the so-called primacy and recency effects~\cite{Robinson,glanzercunitz}, i.e. better recall for items respectively from the beginning and end of a list to memorize as well as enhancement of recall due to adding elaborations~\cite{BowerClark,SteinBransford}. Forgetting in humans occurs primarily through interference rather than decay of memory traces~\cite{JenkinsD}-\cite{Oberauer}. Repetitions work best after some delay~\cite{Dempster}. 
All these properties could be thought of being tied to the specific biological implementation of human memory and being largely independent of the essence of language use. Hence it is very illuminating to study these phenomena in a wholly artificial language system -- a~Large Language Model.

It is important to contrast this paper with the very interesting investigations of higher cognitive properties, such as decision making, causal reasoning or creativity performed for \texttt{ChatGPT} (and other chatbots) or \texttt{GPT-3/4} (see e.g.~\cite{sparks}-\cite{creativity}), where the main interest rests squarely on the AI side. Here we focus instead on a very low-level human cognitive faculty -- (short-term) memory and its specific characteristics observed in humans, which appear inessential for the overall functional role of memory in the context of language. In the course of this study, we effectively end up arriving at some unexpected
interrelationships of humans as biological entities and language, with the LLM used primarily as a sophisticated tool to explore nontrivial statistical properties of language use, as we describe below.

The key result of this paper is that Large Language Models exhibit properties of memory\footnote{Note that in this paper we study quite distinct phenomena from the memorization of passages from training data by some LLMs~\cite{memorization}.} qualitatively similar to the ones characteristic of humans (e.g. see Fig.~\ref{fig.human} for an example of primacy and recency effects). 
In the present paper, the LLM's memory is understood \emph{functionally} and is defined in terms of its probabilistic model of language use, rather than as a \emph{physical subsystem} of the LLM. In fact, the transformer architecture~\cite{transformer} of the GPT-class models does not have any dedicated memory subsystem. Therefore memory appears in these LLMs as an \emph{emergent} phenomenon.

The similarity of the characteristics of human biological memory to LLM's memory can be \textit{a-priori} interpreted in two ways: 
\begin{enumerate}
\item It might be due to the fact that the architectural features of LLMs somehow resemble the workings of human memory.
\item It might be due to the fact that we structure our narratives in a way compatible with the characteristics of our biological memory. 
And the LLMs, trained on these texts, incorporate these subtle statistical imprints into their probabilistic model of language use.
%And the LLMs provide a sufficiently precise and subtle probabilistic model of language use which allows us to detect that.
\end{enumerate}
Either of the two possibilities would be very interesting. In fact, they are not mutually exclusive and there may be some overlap between them.
Later in the paper, however, we provide arguments favouring the latter option.

%Note that the LLM \emph{memory} is understood functionally as some mechanism of retaining semantic consistency, defined in terms of the probabilistic model, rather than as a \emph{physical subsystem} of the LLM. The transformer architecture of the GPT class model does not have any dedicated provision for memory. Therefore memory appears here as an \emph{emergent} phenomenon.

\section{Large Language Models}

The advent of Large Language Models, i.e. generative artificial neural networks serving as underlying models for AI chatbots such as \texttt{ChatGPT}~\cite{chatgpt}, brought about a breakthrough in our ability to model language and generate human-level textual output. These models have been trained on immense corpora of text, effectively building up a very complex and precise probabilistic model of language use.
As chatbots have been in addition specially fine-tuned for conversations through so-called  reinforcement learning from human feedback~\cite{rlhf1,rlhf2}, here we concentrate on LLMs which have been trained 
in the conventional manner
purely on corpora of text and which predict the next word/token\footnote{The relation between tokens and words depends on the tokenizer. For \texttt{GPT-J}, investigated in the present paper, most common words and punctuation marks are single tokens.} based on the preceding text. The trained LLM thus provides for us the conditional probability
\eq
P(\text{token} | \text{preceding text})
\label{e.probmodel}
\eqx
which describes the statistical properties\footnote{This function clearly also has some inductive bias due to the internal architecture of the LLM. We will discuss this point at length later in the paper.} of the corpora of text used for training -- for the LLMs considered in this paper, this is the 825Gb open-source \texttt{Pile} dataset~\cite{Pile}.

Such a language model can be used for text generation by implementing a sampling strategy from (\ref{e.probmodel}) and iteratively building up the generated text. For our purposes, however, we do not need to explicitly generate text and we directly investigate the probabilistic model (\ref{e.probmodel}).

It is important to note that the probabilistic model of language (\ref{e.probmodel}) encompasses a much richer range of phenomena than is commonly considered when thinking about linguistics.
The conditional probabilities include not only constraints on grammatical correctness, but also incorporate an immense amount of factual knowledge 
(which can be objectively correct or not) 
present in the training data. 
All this information is encoded in the weights of the neural network layers\footnote{This could be understood as an analog of human long-term memory.} as a result of training on huge corpora of text. 
%So by construction factual knowledge and grammar 

Even more so, thinking of (\ref{e.probmodel}) as a very general formalization of language use, 
it also has to necessarily incorporate phenomena of a more global nature,
%it is interesting to look at it more globally, 
beyond the scale of grammatical correctness or atomic facts.
Indeed, in order for the language model to be able to generate a coherent narrative, the probabilistic model (\ref{e.probmodel}) has to capture, in addition, long-range semantic correlations within a text. In particular, 
%as already indicated earlier, 
a prerequisite for generating sensible text is having some form of \emph{effective memory} which 
%keeps 
is undestood \emph{functionally} as keeping
track of what has been already stated.
That being said, we note that it is not clear \textit{a-priori} whether this requires the specific memory of atomic facts given in the provided preceding text,  or whether just the general gist would suffice.

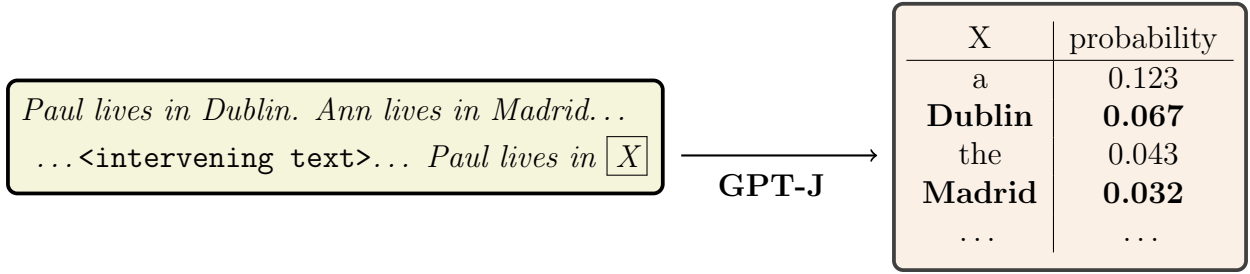
\begin{figure*}[t]

%\hspace{2cm}...Paul lives in \fbox{X} \tikzmark{a}
\tcbox[colback=Beige,colframe=black,right=-8pt]{\parbox{0.52\textwidth}{\it
Paul lives in Dublin. Ann lives in Madrid\ldots\\

\vspace{-0.4cm}

\hspace{0.2cm}\ldots{\tt <intervening text>}\ldots\ Paul lives in \fbox{X} \tikzmark{a}}}
\hspace{2.8cm} 
\tcbox[colback=Linen,right=-2pt]{
\begin{tabular}{c|c}
    X & probability \\
   \hline
   a & 0.123 \\
   \textbf{Dublin} & \textbf{0.067} \\
   the & 0.043 \\
   \textbf{Madrid} & \textbf{0.032} \\
   \ldots & \ldots
\end{tabular}
}

\begin{tikzpicture}[overlay,remember picture, shorten >=-3pt]
\draw[->,thick,rounded corners=8pt] (pic cs:a)+(0.3, 0.15) -- +(2.8, 0.15) ;
\draw (pic cs:a)+(1.5, -0.25) node {\textbf{GPT-J}};
\end{tikzpicture}

\caption{Probing memory in \texttt{GPT-J}. The list of facts is separated from the query by some intervening text.
The large language model \texttt{GPT-J} computes the probabilities of tokens which could be put instead of the X placeholder in the query. We take into account only the tokens corresponding to nouns. The answer is judged as correct if the highest ranking \emph{noun} is identical to the one given for \textit{Paul} in the list of facts.}
\label{fig.overview}
\end{figure*}

This question is especially intriguing, as the LLM neural network architectures considered here do not have  any special memory subsystems. 
Therefore memory, if present, has to
arise in an \emph{emergent} way from the nature of language processing in the transformer sub-units of the Large Language Models.
Even more so, it is important to emphasize that the LLMs have simultaneous access to the \emph{whole} preceding text when computing~(\ref{e.probmodel}).
If memory can at all be identified in them, it has to
be understood only in a \emph{functional} way, 
e.g. as described in the following section.
%In the following section
The goal of this paper is to identify and explore the features of the memory characteristics of Large Language Models and compare them to some aspects of human memory.

%The conditional probability (\ref{e.probmodel}) incorporates in particular i) grammatical constraints, ii) factual knowledge (like \textit{Paris is the capital of France}) and, most importantly for our purposes, iii) various global constraints on narratives, such as e.g. overall semantic consistency.
%Let us note that from the perspective of the method of training of the LLM, all these ingredients appear on the same footing, thus for the LLM factual knowledge is an intrinsic part  
%In particular, the generated text should not contradict itself. Therefore, the LLM should retain knowledge of facts provided in the preceding text, when generating subsequent parts of the narrative through~(\ref{e.probmodel}). This means that some form of \emph{memory} of transient facts given in the preceding text should be present in the LLM.
%Note that this \emph{memory} is understood functionally as some mechanism of retaining semantic consistency, rather than as a \emph{physical subsystem} of the LLM. The transformer architecture of the GPT class model does not have any dedicated provision for memory. Therefore memory appears here as an \emph{emergent} phenomenon.
%The goal of this paper is to study the properties of this emergent LLM memory.

\section{Probing memory in Large Language Models}
\label{s.probingmemory}

The standard serial memory test paradigm in cognitive psychology consists of giving a subject sequentially a list of words, and then testing recall accuracy as a function of the position on the list. 

It is not entirely obvious how to construct an analog of such a human memory experiment for a Large Language Model, as one should not rely on providing instructions for the memory task to the neural network -- instructions, which are of course given to a human participant. 
Although one could envisage doing that for chatbots, the interpretation of the results would then be less clear, as it would in addition be confounded by testing the comprehension of the %specific wording of the 
instructions by the chatbot. In case of chatbots, one would also face having to understand the impact of the more involved (and largely undisclosed) method of training chatbots with human instructors. 
Therefore, we restrict ourselves to conventional LLMs and we leverage instead the conditional probability model (\ref{e.probmodel}) by specifically constructing the \textit{preceding text} to probe a particular memory phenomenon.

Instead of memorizing a list of words, the network is presented with a list of elementary facts concerning a set of arbitrary persons identified by their first names.
We primarily consider facts with a \textit{has-a} relationship:

\exampletext{Paul has a guitar. Ann has a bike...}

\noindent{}In addition, we also investigated an \textit{is-a} relationship:

\exampletext{Paul is a physicist. Ann is a programmer...}

\noindent{}and facts of a \textit{lives-in} type:

\exampletext{Paul lives in Dublin. Ann lives in Madrid...}

\noindent{}After a further intervening text\footnote{Unless stated otherwise, as the intervening text we use \textit{Now, after you received all this information, try to concentrate, drink a cup of coffee, go for a walk. Then please complete the following sentence.} 
%Later in the paper, we modify the intervening text to probe various memory phenomena.
%We also investigate some effects of modifying the intervening text.
}, we append a query such as

\bigskip
\noindent{}\textit{Paul lives in \fbox{X}}\\

\noindent{}and the language model provides probabilities for all tokens in place of $X$. 
%In order to interpret the results as a test of memory, one has to be cautious, as the generative language model may ``want'' to continue the given text e.g. like \textit{Paul lives in a beautiful town...}. Therefore we 
We
restrict ourselves only to tokens representing nouns (including proper names and numbers) and consider as the correct answer the situation when the probability 
\eq
P(\text{Dublin} | \text{preceding text})
\eqx 
is largest among all nouns. This process is illustrated in Fig.~\ref{fig.overview}.

When choosing concrete examples of the elementary facts,
we have to ensure that the correct answer is a word represented by a single token and, in the case of \textit{has-a} or \textit{is-a} relationships, we ensure that it starts with a consonant as a change of the article from \textit{a} to \textit{an} would provide an unwanted hint for the correct answer.

We use a list of 20 names and 20 words from each category (objects, place names and occupations). 
We construct various experiments by modifying the length of the list of facts to be memorized, the category and the intervening text in order to probe various memory phenomena.
We obtain 30 repetitions of each experiment by randomly permuting both names and target words and retaining the appropriate number of facts to construct a list of given length. This is done so that any more ``memorable'' words for the LLM would not bias the results.
When reporting recall accuracy for a given position in the list, we repeat the above experiment with~5 random seeds in order to increase statistics, giving in total $5\times 30 =150$ repetitions.
We provide full details of the experiments in the \textit{Supplementary Information}.
Throughout the paper we primarily investigate the open source \texttt{GPT-J} model with around 6 billion parameters~\cite{gptj}. In one set of experiments, in order to study dependence on model size, we take a selection of LLMs from the \texttt{Pythia} family~\cite{Pythia}. The code for all experiments is provided at \href{https://github.com/rmldj/memory-llm-paper}{\texttt{github.com/rmldj/memory-llm-paper}}.

\section{Results}

In this section we review the results of our experiments probing the LLM analogs of the properties of human memory: primacy and recency effects, the influence of elaborations on memory recall, forgetting as memory decay or interference; we also study the impact of repetitions. In addition, we discuss a LLM-specific feature -- ``memory formation time'' -- which will be important for interpreting the overall results.

\subsection*{Primacy and recency effects}

A very characteristic feature of human memory when memorizing lists of words is the fact that words from the beginning and from the end of the list are easier to recall, phenomena called \emph{primacy} and \emph{recency effects}~\cite{glanzercunitz} (see sample human data in Fig.~\ref{fig.human}).
In order to investigate similar properties for Large Language Models, we calculate the accuracy of recall as a function of position of the given fact in the list of facts.

\begin{figure}[t!]
\begin{center}
\includegraphics[width=0.45\textwidth]{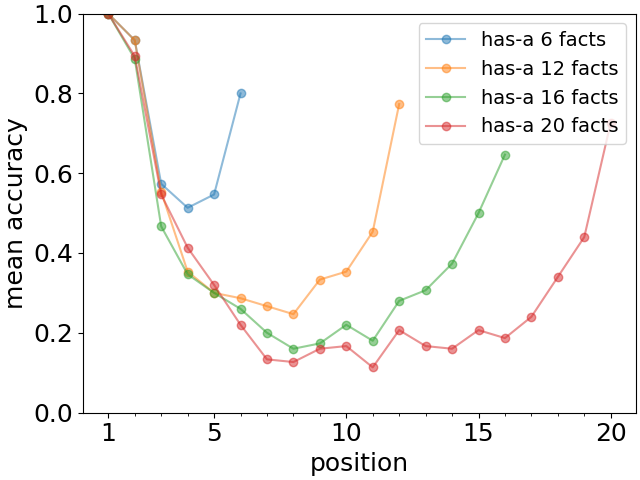}\\
\includegraphics[width=0.45\textwidth]{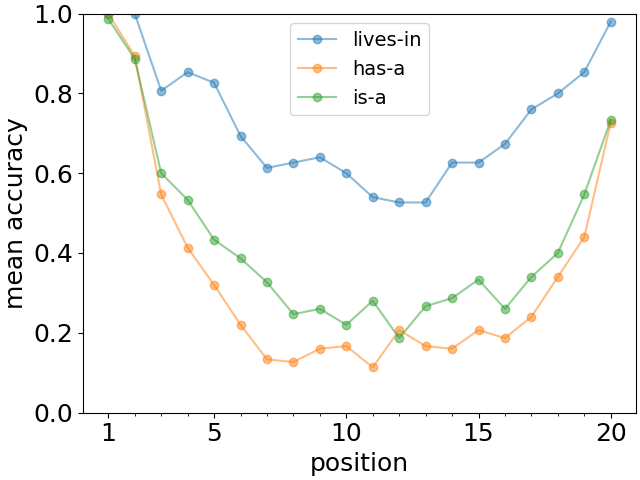}
\end{center}
  \caption{Recall accuracy as a function of position for lists of various lengths (top) and for facts of various types (bottom).}
  \label{fig.overallbis}
\end{figure}

The results for various lengths of the list are shown in Fig.~\ref{fig.overallbis} (top). We observe a very clear pattern characteristic of the \emph{primacy} and \emph{recency effects}. 
Moreover, the recall accuracy for the first two or three facts remains essentially unchanged with the length of the list. Similarly, the recall of the final one or two facts is quite stable.  
In Fig.~\ref{fig.overallbis} (bottom) we show for lists of length 20 that the overall qualitative U-shaped recall pattern occurs for all three relationships. Note that even though the \textit{lives-in} case appears to be easier to remember for the LLM, the primacy and recency effects are still clearly present.

\begin{figure}[t]
\begin{center}
\includegraphics[width=0.45\textwidth]{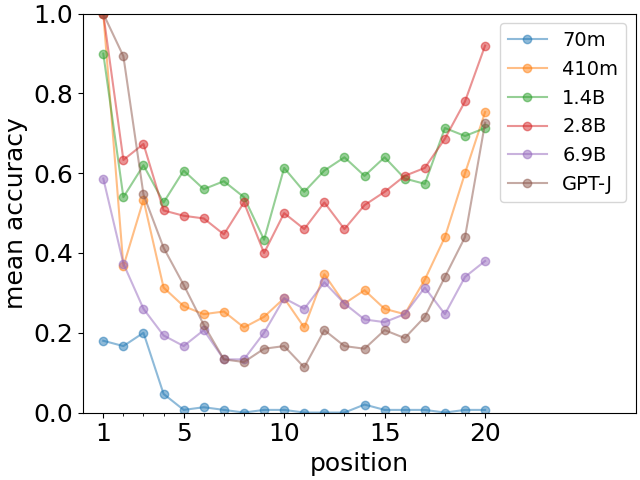}
\end{center}
  \caption{Recall accuracy as a function of position for a list of 20 facts with a \textit{has-a} relationship for a variety of \texttt{Pythia}-family language models of different sizes. 
  %from the \texttt{Pythia} family.
  }
  \label{fig.pospythia}
\end{figure}

In order to check the robustness of the U-shaped recall curve, we also performed identical experiments\footnote{The Pythia tokenizer is, however, different and less convenient for these experiments. See \textit{Supplementary Information} for more details.} with a range of LLMs of the \texttt{Pythia} family~\cite{Pythia} differing widely in the size of the model. The results are shown in Fig.~\ref{fig.pospythia} for the updated models as of April 3, 2023.
%We observe that the clear U-shaped recall curve gets more and more visible with increasing  model size. Interestingly, the smallest model (\texttt{Pythia-70m} with just around $1/100$ of the number of parameters of \texttt{GPT-J}) exhibits only the primacy effect. For larger models, recency effect also appears, with the smoothest curve, especially in the part relevant for the primacy effect, appearing for the largest models.
We observe that the primacy effect is clearly visible for all models, while the recency effect appears for a subset. Of particular note is the complete absence of the recency effect for the smallest model (\texttt{Pythia-70m} with just around $1/100$ of the number of parameters of \texttt{GPT-J}). In \textit{Supplementary Information} Fig.~S1, we show an analogous plot for the previous versions of the \texttt{Pythia} models, where the training regimen was dependent on the model size. In that case, recency effect appears much more systematically (but also apart from the smallest model) and the U-shaped recall curve is clearly visible.

To summarize, we observe that the primacy and recency effects are a very generic feature of Large Language Models requiring, however, that the model be large enough for their full development.
The primacy effect seems very stable, while the recency effect may be more brittle and sensitive to peculiarities of training as can be seen comparing the two versions of the \texttt{Pythia} models.
When both effects are present, as for the \texttt{GPT-J} model studied throughout this paper, they appear systematically irrespective of the type of facts or list length as in~Fig.~\ref{fig.overallbis}.

\subsection*{Elaborations and recall}

Another distinctive feature observed in human memory tests is that providing additional information about a given concept improves the chances of its recall, even if the query does not involve any of the additional provided information~\cite{BowerClark,SteinBransford}. In order to test whether a~similar phenomenon occurs also for Large Language Models, we consider a baseline list of length 19 of facts from the \textit{has-a} relationship. Then at positions 5, 10 and 15 in the list, we add elaborations on these facts.~E.g.

\exampletext{Paul has a guitar.}

\noindent{}in the baseline list is substituted with

\exampletext{Paul has a guitar, an electric one, on which he plays in a local garage band.}

\begin{figure}[t!]
\begin{center}
\includegraphics[width=0.45\textwidth]{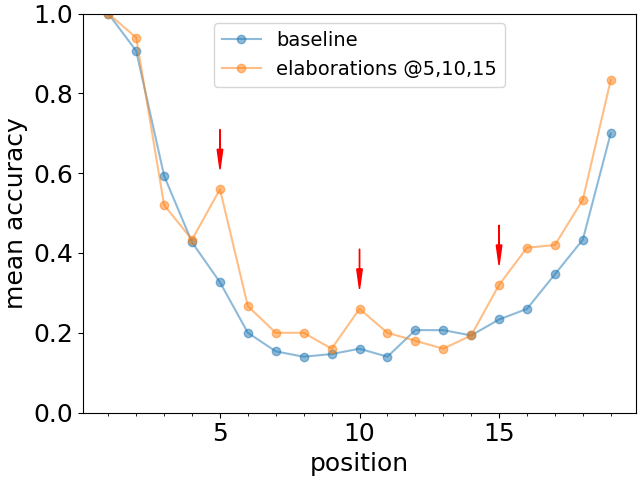}
\end{center}
  \caption{Comparison of recall for a baseline list of facts and with elaborations added at positions marked with red arrows.}
  \label{fig.elaborations}
\end{figure}

\noindent{}Then we again query just for the \textit{has-a} relationship, thus not invoking any of the additional knowledge contained in the elaboration.
The comparison of the recall accuracy as a function of position for the baseline list and the list with the elaborations added at the selected positions is shown in Fig.~\ref{fig.elaborations}. We observe that elaborations indeed clearly increase the recall accuracy, effectively increasing the perceived saliency of a given fact.
%Adding elaborations effectively increases the saliency of a given fact.
The full list of elaborations is listed in the \textit{Supplementary Information}.

\subsection*{Interference and forgetting}

\begin{figure}[t!]
\begin{center}
\includegraphics[width=0.45\textwidth]{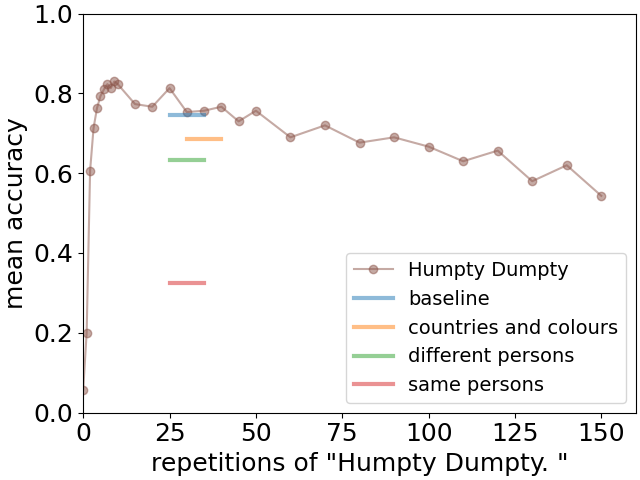}
\end{center}
  \caption{Performance of recall as a function of the number of repetitions of \textit{Humpty Dumpty} between the list of 10 facts and the query. The horizontal bars indicate performance with interfering information provided instead. The bars are located at the approximate equivalent length of the full intervening text.}
  \label{fig.decay}
\end{figure}

Forgetting, or the loss of memories, can be \emph{a-priori} associated with two quite distinct mechanisms. One possibility would be that the traces of memory would fade, and thus the main mechanism would be \emph{memory decay}. Alternatively, new memories would overwrite old ones, and the main mechanism of forgetting would be through \emph{memory interference}. 
Although it is certainly true that both mechanisms are at play in forgetting, a number of psychological studies~\cite{JenkinsD}-\cite{Oberauer} identified \emph{memory interference} as the dominant factor.

In order to construct an investigation of analogous phenomena on the LLM side, we will now change the \emph{intervening text} (see Fig.~\ref{fig.overview}) in an appropriate way to model each of those mechanisms.

For investigating \emph{memory decay}, we use intervening text which basically does not carry any semantic content for any length.
Concretely, we take $n$ repetitions of \textit{``Humpty Dumpty.''}, for which we use a shorthand notation
\[
n \times HD \equiv n \times \text{\it ``Humpty Dumpty. "}
\]
The results of the mean recall accuracy as a~function of $n$ are shown as the solid curve with dots in Fig.~\ref{fig.decay}. We observe a slow decline in the performance with increasing\footnote{Here we focus on the dependence for moderate to large $n$. We will discuss the initial rise for $n<10$ in the subsection on \textit{Memory formation time}.} $n$, showing that an analog of \emph{memory decay} can be identified in the LLM context. What is crucial, however, is to compare this with the decline due to interference.

\begin{figure}[t]
\begin{center}
\includegraphics[width=0.45\textwidth]{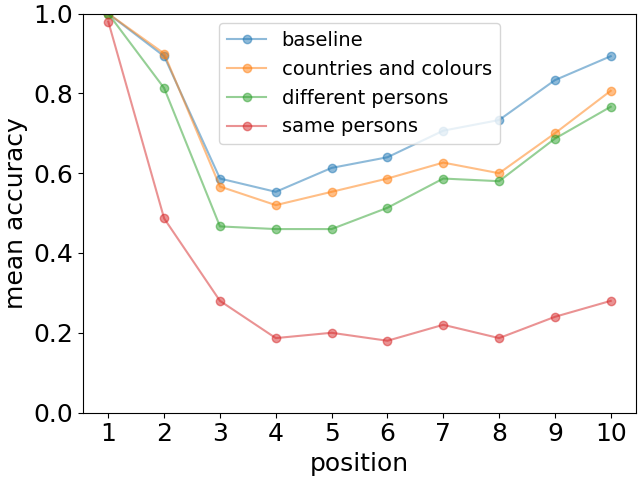}
\end{center}
  \caption{Performance as a function of list position for various kinds of interfering information given in the intervening period.}
  \label{fig.interference}
\end{figure}

For modelling interference, we take the intervening text to be
\[
10\times HD + \text{\bf distractor} + 10\times HD
\]
where the \textbf{distractor} is a list of 10 distracting facts of various kinds: i) colours of countries on a map, ii) occupations of 10 persons \emph{different} from the ones given in the original list of facts, and iii) occupations of the \emph{same} 10 persons mentioned in the original list.
Note that none of these distracting facts are in contradiction with the original ones. They are either completely independent (as in cases i) and ii)) or provide complementary information (case iii)) e.g. \textit{Paul is a physicist} in addition to \textit{Paul has a guitar}.
The recall performance is indicated in Fig.~\ref{fig.decay} by horizontal bars. The baseline is when the distractor is replaced just by~10 repetitions of \textit{Humpty Dumpty.}.

We observe that interference decreases recall accuracy much stronger than decay, with the decrease depending on the type of interfering information. This is especially evident for distracting facts involving the \emph{same} set of persons as in the original list. In this case we also observe the disappearance of the \emph{recency effect} (see Fig.~\ref{fig.interference}).

\subsection*{Repetitions}

Repetition of a given material clearly should increase its memorization. We expect that LLMs should behave similarly in this respect. In case of human memory, since the time of Ebbinghaus it has been well established that repetition works best if it is separated by a certain interval of time from the initial presentation of the material to be learned~\cite{Dempster}.
There does not seem to be any obvious reason for this to hold for LLMs also, as this property should rather be tied to features of biological memory consolidation.

\begin{figure}[t]
\begin{center}
\includegraphics[width=0.45\textwidth]{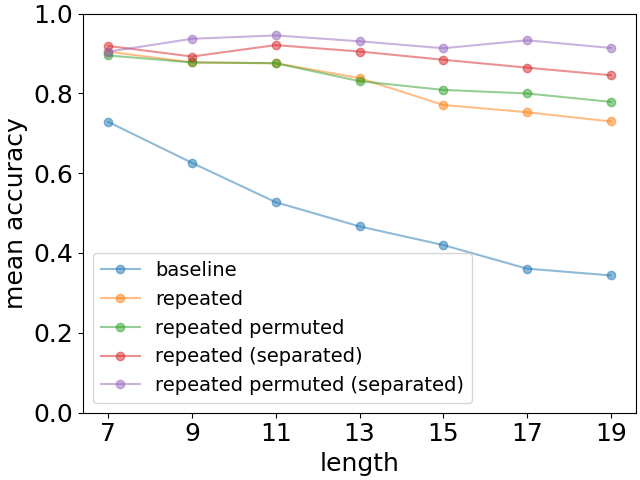}
\end{center}
  \caption{Performance improvement after repeating the given facts in the intervening period. Introducing a separation period prior to repetition improves recall.}
  \label{fig.repetitions}
\end{figure}

In order to investigate the memory recall performance of the LLM with repetitions, we precede the standard intervening text of section~\ref{s.probingmemory} with either
\eq
\text{\bf repetition} + 10\times HD
\eqx
or
\eq
10\times HD + \text{\bf repetition}
\label{e.separated}
\eqx
where \textbf{repetition} is the repeated list of facts to memorize. We optionally also permute the ordering of facts in the repeated list. As a baseline, we precede the standard intervening text with an appropriate number of insertions of \textit{Humpty Dumpty} to match the length of the repetition:
\eq
(10 + n) \times HD
\eqx
The results are shown in Fig.~\ref{fig.repetitions}. As expected, repetitions significantly improve recall accuracy w.r.t. the baseline. Moreover, we observe a clear improvement of performance when the repetition is \emph{separated} from the original list of facts as in \eqref{e.separated}, which is in line with what happens for humans~\cite{Dempster}.
In the latter case, we also see that the permuted case has slightly better accuracy\footnote{The difference is quite small, but it persists systematically for \emph{all} list lengths greater than 7. If it would be just an effect of statistical fluctuations, we would expect the difference to have varying signs for various list lengths.}. 
We are unaware if the latter phenomenon has been investigated for human subjects.

\subsection*{Memory formation time}

We close the presentation of the obtained results with a phenomenon which seems specific to LLMs, and which is not expected to occur for human subjects. In Fig.~\ref{fig.decay}, we observe that the performance initially strongly increases from very low levels, reaching a maximum for the intervening text being around 10 repetitions of \textit{Humpty Dumpty}.
It therefore appears that there is some effective ``memory formation time'' necessary for prior information to be available for use later in the text.
This behaviour is quite counterintuitive, as one would expect that the recall of facts should be easiest directly after the facts are given. This runs apparently also against the clear and very robust observation of the \textit{recency effect} described earlier for LLMs. These observations will be important for discussing the possible interpretations of the results obtained in the present paper.

\section{Discussion}

In this paper we have demonstrated that a Large Language Model -- \texttt{GPT-J} -- exhibits a number of key qualitative characteristics of human memory: i) primacy and recency effects, ii) improvement of recall due to added elaborations, iii) forgetting primarily through interference rather than memory decay and iv) the benefit of repetitions separated by some interval.
Here LLM's memory was defined in terms of the conditional probability~\eqref{e.probmodel} through an appropriate construction of the preceding text fed to the network. This amounts to a \emph{functional} definition of memory, with the LLM acting \emph{as if} it were to participate in a serial memory test.

Such a close similarity of the characteristics of human and LLM memory is in fact very surprising and begs explanation.
This question is especially intriguing as the LLMs of the type considered in the present paper do not have any dedicated memory subsystem, and thus their ``memory'' arises in an \emph{emergent} manner. 

The conditional probability~\eqref{e.probmodel}, being at the root of the definition of LLM memory, is wholly dependent on two factors: i) the specific neural network architecture of the Large Language Model~\cite{llm,transformer,gptj} and ii) the training text corpora -- here the 925Gb \texttt{Pile} dataset~\cite{Pile}. Depending on the relative importance of these two factors, we may arrive at the two possible interpretations already summarized at the beginning of this paper.

If the specific features of the LLM architecture would bias the probabilistic model~\eqref{e.probmodel} in a way leading to all these human-like characteristics of memory, we could suspect that the inner workings of the transformer model somehow capture (functionally) some aspects of the working of human memory\footnote{Of course, there would still be a theoretical possibility of the existence of two completely independent inductive biases (human and LLM) leading to the same properties.}.

If, on the other hand, the deep neural network architecture of the Large Language Model would function really as a universal function approximator with enough capacity to model the very complex statistical interrelations in human text corpora, then the architectural bias would be unimportant. Consequently, we would attribute the human-like memory properties to \emph{genuine} statistical correlations in the textual training data, with the LLM being a sufficiently precise tool to detect them.
This would mean that we, humans, structure our texts in a way which is compatible with the properties of our biological memory.

It is \emph{a-priori} quite difficult to distinguish between the two possibilities, especially as there may be, of course, some overlap between the architectural inductive bias and genuine properties of training data. Ideally, a distinction could be made if we had at our disposal a \emph{similarly performant} language model but with a quite different internal architecture. Unfortunately, currently all LLMs of comparable performance are based on transformers, so we have to confine ourselves to indirect arguments.

A crucial difference between the inner workings of the transformer-based LLM and a human when processing text is the \emph{absence} of the factor of the passage of time in the Large Language Model. Indeed, when computing the conditional probability~\eqref{e.probmodel}, the LLM has at its simultaneous disposal the \emph{whole} preceding text. 
Hence, there is no issue in preserving or losing information in time as in real short-term memory of humans.
Internally, the transformer computes (weighted) similarities between tokens (and their representations deeper in the network) at \emph{any} positions in the preceding text -- so also directly between the token in the query and the tokens in the given list of facts. The LLM memory properties amount thus to their capacity to analyze the preceding text and extract relevant information.
The overall mechanism is thus apparently quite different.

Let us now discuss the possible inductive biases of the LLM architecture for the human-like characteristics of memory.
The position of the word/token in the LLM is encoded by adding so-called (rotary) positional embedding~\cite{rotary}, which is effectively the only proxy for time. Consequently, the neural network knows which tokens are far away and which are adjacent. 
Moreover, the overlap of the embedding vectors decreases with the inter-token distance.
From this perspective, one could think that this provides a very clear architectural bias for the \emph{recency} effect. We can make two arguments, however, that this is not the case. 

Firstly, the smallest \texttt{Pythia-70m} model, which has the same kind of positional embedding, does not exhibit the recency effect at all (see Fig.~\ref{fig.pospythia}). 
The effect only develops for larger models, being also more brittle with respect to the training regimen, indicating that it is \emph{learned} and not just induced by an architectural bias.

Secondly, the effect of ``memory formation time'' exhibits initially rising accuracy with \emph{increasing} distance of the query from the list (see the continuous curve on Fig.~\ref{fig.decay}). This is again counter to the architectural bias hypothesis of recency, which would imply an opposite effect. 
%However, even in this case, for a given length of the intervening text, recency effect is still present.

The only characteristic of memory which might well be directly influenced by the architecture of the Large Language Model is forgetting through interference.
As the main mechanism of the transformer is essentially (weighted) similarity matching, interfering facts associated to the same persons may well strongly decrease recall (see Fig.~\ref{fig.decay}) as they could interfere with the similarities. Indeed, both facts associated to the same person could be simultaneously selected during the matching procedure and hence could interfere with each other in the process of ``answering'' the query. 

From the above considerations we conclude that it is much more probable to attribute the major part of the similarities of the characteristics of human and LLM memory to genuine properties of the textual data used to train the LLM, and thus to human textual output.
This indicates that semantic correlations characteristic of the qualitative features of human memory are somehow present in the training texts. 
We thus arrive at the conclusion that the properties of our biological memory leave an imprint on how we structure globally our textual narratives. This fascinating possibility indicates that language and biology are more intertwined than one would naively expect.
We hope that the above perspective would engender further research along these lines. 

\bigskip

\noindent{\bf Acknowledgements.} 
I would like to thank Tadeusz Marek, Magdalena Fafrowicz, Igor Podolak, Natasha Klein-Atlas for discussions and comments on the manuscript, and Michael Olesik for collaboration in the initial stage of this project.
 This work was supported by the research project \textit{Bio-inspired artificial neural networks} (grant no. POIR.04.04.00-00-14DE/18-00) within the Team-Net program of the Foundation for Polish Science co-financed by the European Union under the European Regional Development Fund and by a grant from the Priority Research Area DigiWorld under the Strategic Programme Excellence Initiative at Jagiellonian University.

\clearpage
\onecolumn

\renewcommand{\thefigure}{S\arabic{figure}}

\section*{Supplementary information}

\subsubsection*{Details on the experimental tasks}

In the experiments performed in the paper, we consider 20 persons identified by the following first names:

\bigskip
\noindent{}\textit{Paul, Helen, Ann, Mary, David, Mark,
                Michael, Susan, Robert, Peter, Christine, Sarah,
                Ivan, Charlotte, Pierre, Catherine,
                Audrey, John, Amanda, Kevin
}

\bigskip

\noindent{}For the \textit{has-a} relationship primarily studied in the paper, the facts are of the form \textit{N has a X}, where $N$ is one of the names and $X$ is one of

\bigskip
\noindent{}\textit{bike, cat, dog, guitar, piano,
                camera, laptop, motorcycle, house, sister, brother,
                trumpet, keyboard, violin, Toyota, Porsche,
                Ford, Mercedes, horse, boat
}

\bigskip

\noindent{}For the \textit{is-a} relationship, the facts are of the form \textit{N is a X}, where $X$ is one of

\bigskip
\noindent{}\textit{biologist, driver, farmer, mathematician, physicist, programmer,
                journalist, lawyer, doctor, surgeon, psychologist, politician,
                nurse, teacher, writer, soldier, 
                pilot, baker, painter, musician
}

\bigskip

\noindent{}For the \textit{lives-in} relationship, the facts are of the form \textit{N lives in X}, where $X$ is one of

\bigskip

\noindent{}\textit{Dublin, Copenhagen, Budapest, Warsaw, Madrid, Stockholm,
                   Tokyo, Sydney, Delhi, Seattle, Havana, Cairo,
                   Melbourne, Chicago, Lisbon, Honolulu,
                   Seoul, Rome, Athens, Manila
}

\bigskip

\noindent{}The elaborations used for the particular objects for the \textit{has-a} relationship were chosen as follows:

\bigskip
{\it\noindent{}N has a bike, on which he/she drives to work each day.\\
N has a cat, which passionately likes to play with a ball.\\
N has a dog, called Fido who just adores catching his rope toy.\\
N has a guitar, an electric one, on which he/she plays in a local garage band.\\
N has a piano, which unfortunately is a bit out of tune.\\
N has a camera, a quite heavy full-frame digital SLR with a couple of lenses.\\
N has a laptop, which is covered with all kinds of stickers.\\
N has a motorcycle, not a Harley-Davidson but rather an unasuming model which easily blends in.\\
N has a house, situated in large garden in a nice quiet part of the town.\\
N has a sister, who is much younger, so they did not overlap at high school.\\
N has a brother, who went to school one year earlier, so school was a familiar ground.\\
N has a trumpet, on which he/she regularly plays each weekend in the local jazz club.\\
N has a keyboard, on which he/she tries to practice reading notes and playing standards.\\
N has a violin, on which he/she tries to practice every morning to the dismay of neighbors.\\
N has a Toyota, an old model bought quite cheaply at a second-hand outlet.\\
N has a Porsche, painted red, almost matching the stereotypical image.\\
N has a Ford, a sturdy pickup truck, very useful for the gardening business.\\
N has a Mercedes, which so far has been very reliable, but now some minor problems start to surface.\\
N has a horse, which is kept on a farm just a few miles north of the city.\\
N has a boat, really a small dinghy, used for fishing on the lake.
}

\smallskip
\noindent{}where \textit{he/she} stands for the appropriate pronoun for the given name $N$.

\bigskip
\noindent{}The list of facts is separated from the query by an intervening text.
Unless stated otherwise, we take it to be

\bigskip

\noindent{}\textit{Now, after you received all this information, try to concentrate, drink a cup of coffee, go for a walk. Then please complete the following sentence.}

\bigskip
\noindent{}The interfering facts referred to as \textit{countries and colours} in Fig.~6 and 7, are as follows:

\bigskip
{\it\noindent{}The color of France on the map is blue\\ 
The color of Finland on the map is white\\
The color of Spain on the map is yellow\\
The color of Japan on the map is purple\\
The color of Italy on the map is green\\
The color of India on the map is brown\\
The color of Greece on the map is violet\\
The color of Brazil on the map is orange\\
The color of Denmark on the map is gray\\
The color of Mexico on the map is red. 
}

\subsubsection*{Technical details}

In order to ensure that none of the names or objects biases the results, the names and objects were independently permuted 30 times and the appropriate number of facts was selected. In all cases all names and objects/occupations/places were distinct. In case of measuring the accuracy as a function of position on the list, these experiments were repeated 5 times with 5 different random seeds providing in total 150 repetitions. Consequently, for each position in the list we get 150 binary answers (true/false recall). This data is wholly characterized by specifying the mean accuracy at that position in the list. For such binary data, the accuracy is already by itself a sufficient statistic.
The neural network employed for the vast majority of experiments was \texttt{GPT-J}, evaluated on an NVIDIA V100 GPU.

\subsubsection*{Experiments with Pythia LLMs}

The experiments involving the Pythia family of networks, summarized in Fig.~4 in the main text, required some care. Due to a different tokenizer, many words which were made up of a single token for \texttt{GPT-J} became composite (this occurred for \textit{trump-et} and \textit{P-orsche} in the words relevant for the \textit{has-a} relationship). In such a situation only the first token was tested.
Also in this case we did not restrict ourselves to tokens which were nouns due to the fact that the correspondence between words and tokens seemed to be much looser than for \texttt{GPT-J}. 

\subsubsection*{Code repository}

All code used in performing the experiments in this paper is available at \href{https://github.com/rmldj/memory-llm-paper}{\texttt{github.com/rmldj/memory-llm-paper}}.

%\pagebreak

\vspace{3cm}

\begin{figure}[h]
\begin{center}
\includegraphics[width=0.75\textwidth]{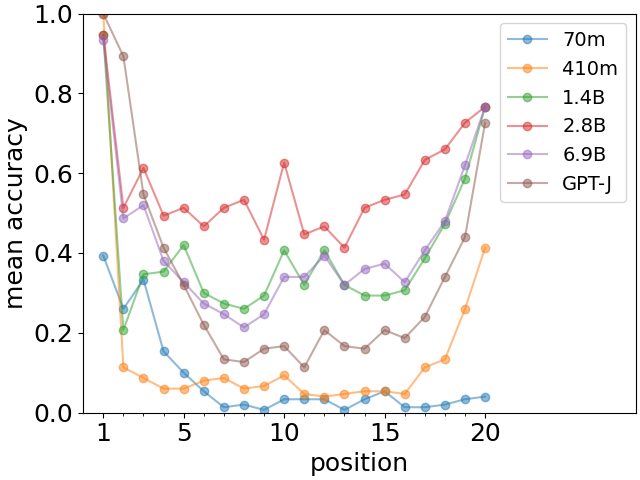}
\end{center}
  \caption{Recall accuracy as a function of position for a list of 20 facts with a \textit{has-a} relationship for a variety of \texttt{Pythia}-family language models of different sizes. These are versions prior to the ones presented in Fig.~4 in the main text, whose training regimen was slightly different for various model sizes.
  %from the \texttt{Pythia} family.
  }
  %\label{fig.pospythia}
\end{figure}

\end{document}